\documentclass{article}

\usepackage[final]{neurips_2024}

\usepackage[utf8]{inputenc} 
\usepackage[T1]{fontenc}    
\usepackage[utf8]{inputenc}
\usepackage{hyperref}
\usepackage{url}            
\usepackage{booktabs}       
\usepackage{amsfonts}       
\usepackage{nicefrac}       
\usepackage{microtype}      
\usepackage{xcolor}         
\usepackage{times}
\usepackage{soul}
\usepackage{url}
\usepackage{graphicx}
\usepackage{amsmath}
\usepackage{amsthm}
\usepackage{booktabs}
\usepackage{algorithm}
\usepackage{algorithmic}
\usepackage[switch]{lineno}
\usepackage{amssymb}
\newtheorem{theorem}{Theorem}

\newtheorem{lemma}[theorem]{Lemma}

\newtheorem{Problem}{Problem}

\usepackage{comment}

\newtheorem{definition}{Definition}

\title{Adaptive Wizard for Removing Cross-Tier Misconfigurations in Active Directory (Extended Version)}

\author{%
  Huy Q. Ngo\\
  School of CMS\\
  The University of Adelaide\\
  Adelaide, SA 5005 \\
  \texttt{quanghuy.ngo@adelaide.edu.au} \\
  \And
  Mingyu Guo\\
  School of CMS\\
  The University of Adelaide\\
  Adelaide, SA 5005 \\
  \texttt{mingyu.guo@adelaide.edu.au} \\
  \AND
  Hung X. Nguyen\\
  School of CMS\\
  The University of Adelaide\\
  Adelaide, SA 5005 \\
  \texttt{hung.nguyen@adelaide.edu.au} \\
}

\begin{document}

\maketitle

\begin{abstract}
Security vulnerabilities in Windows Active Directory (AD) systems are typically modeled using an attack graph and hardening AD systems involves an iterative workflow: security teams propose an edge to remove, and IT operations teams manually review these fixes before implementing the removal.  As verification requires significant manual effort, we formulate an Adaptive Path Removal Problem to minimize the number of steps in this iterative removal process.  
In our model, a wizard proposes an attack path in each step and presents it as a set of multiple-choice options to the IT admin. The IT admin then selects one edge from the proposed set to remove.
This process continues until the target $t$ is disconnected from source $s$ or the number of proposed paths reaches $B$. The model aims to optimize the human effort by minimizing the expected number of interactions between the IT admin and the security wizard. We first prove that the problem is $\mathcal{\#P}$-hard. We then propose a set of solutions including an exact algorithm, an approximate algorithm, and several scalable heuristics. 
Our best heuristic, called DPR, can operate effectively on larger-scale graphs compared to the exact algorithm and consistently outperforms the approximate algorithm across all graphs.
We verify the effectiveness of our algorithms on several synthetic AD graphs and an AD attack graph collected from a real organization.
\end{abstract}

\maketitle

\section{Introduction}

We propose the Adaptive Path Removal Problem, a model motivated by the challenge of eliminating attack paths in cybersecurity. We begin by describing the cybersecurity use case that motivates our approach and by explaining the design rationale behind our model. The main contributions of this paper are the introduction of a novel theoretical model and the exploration of scalable algorithms for solving this problem. Our model’s design rationale is heavily influenced by practical cybersecurity scenarios and by the urgent demand for workable solutions from security teams.


Windows Active Directory (AD) is Microsoft's directory service that enables IT administrators to manage security permissions and control accesses across Windows domain networks. 
An AD environment is naturally described as a graph where nodes are accounts/computers/groups, and the directed edges represent accesses/permissions/vulnerability.
One of the main focus in this line of work is minimizing “attack paths”—routes an attacker might use to escalate privileges and move laterally within the network.

Existing security models \cite{guo2023scalable,zhang2024practical,goel2023evolving} and commercial tools such as BloodHound \cite{BloodHound} reduce these paths by suggesting actionable fixes, typically presented as sets of edges to remove from the graph.
Unfortunately, not every proposed fix (edge) is implementable. Some edges may appear redundant, but removing them could cause significant disruptions.
Since removing edges equates to revoking permissions or accesses within the network, each fix must be approved and implemented by IT operations teams.
This has been referred to in the literature as the “implementable fixes” problem \cite{dunagan2009heat,guo2024limited}.
In industry practice, network hardening workflow typically unfolds in two stages: the security team first proposes necessary fixes, and then IT operations team review those fixes before implementation. This practical constraint has motivated the development of adaptive security models, models that incorporate human feedback and are thus better suited to real-world usage.

In the same way a “proxy” in auction theory places bids on user’s behalf, our proposed wizard model acts as a “proxy” security operator that guide the IT administrator through the attack path removal process.
At every step, the wizard model proposes an attack path to remove.
The IT admin view this as a multiple-choice list of edges and will require to choose one edge to remove.
This process continues until all attack paths are eliminated, or until the number of proposals reaches a preset limit.
The wizard’s goal is to minimize the expected number of proposals.
The wizard is adaptive, meaning it proposes subsequent edges to remove based on the IT admin’s choices in previous steps. 
Unlike previous work \cite{guo2024limited,zheng2011active,dunagan2009heat}, which modeled IT admin’s decision as simply removing or retaining an edge (binary decision), without guaranteeing that all attack paths would be eliminated; our path-based proposal mechanism provides a cut-guarantee solution.

\begin{figure}[h]
 \centering
  \includegraphics[width=0.5\linewidth]{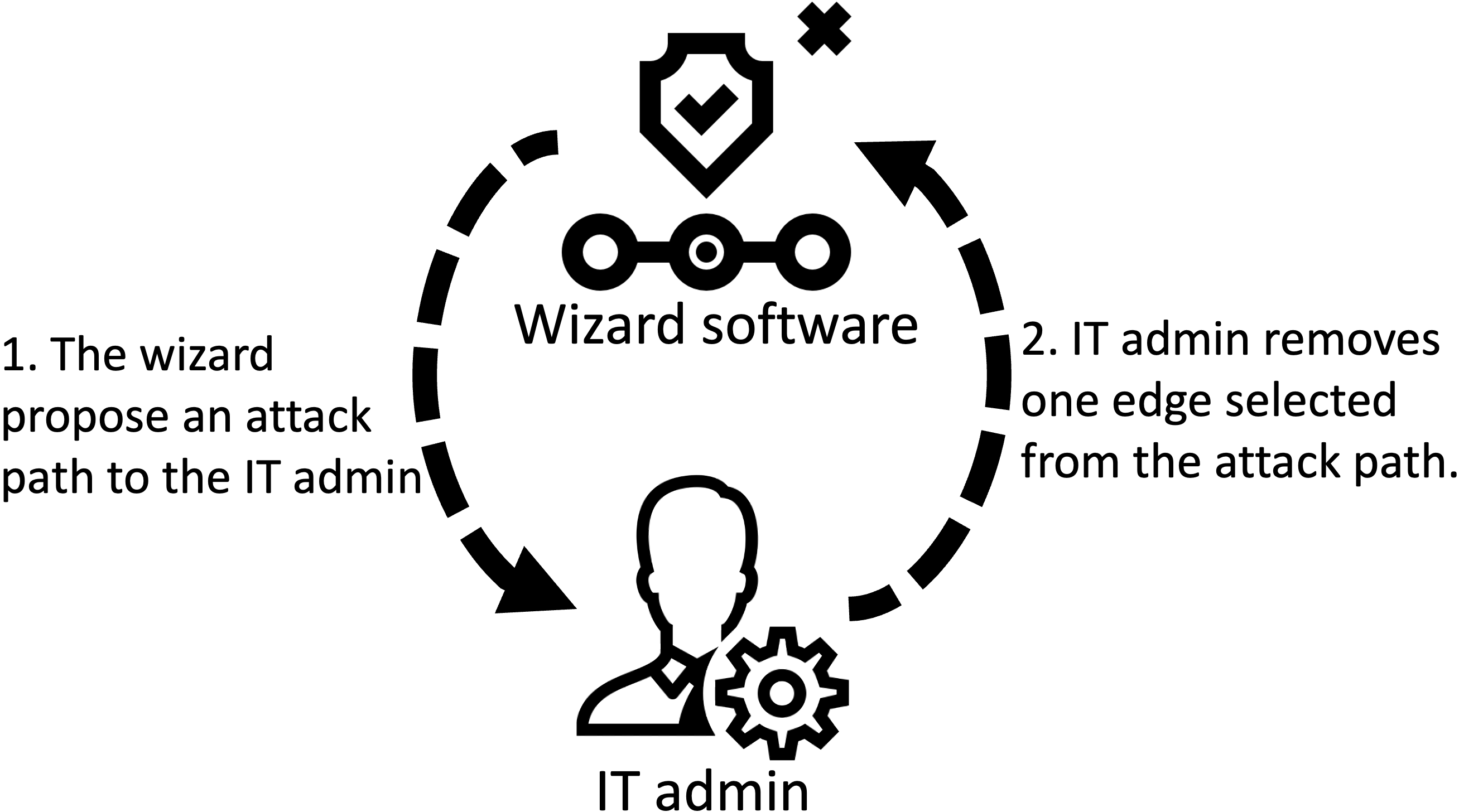}
  \caption{The wizard is a software step-by-step guide to assist the user in performing correction actions without requiring extensive technical knowledge.}
  \label{fig:wizard}
\end{figure}


Our key contributions can be summarized as follows:

\begin{itemize}
    \item We introduce a new theoretical combinatorial optimization model called Adaptive Path Removal, motivated by the network security use case in AD systems. This is the first adaptive graph-focused model to incorporate path proposals and provide a cut-guarantee solution.
    \item We prove that the problem is $\#\mathcal{P}$-hard and introduce both an exact and an approximate algorithm.
    \item We develop a scalable heuristic called Dynamic Programming with Restriction (DPR), which builds on our exact and approximate algorithms. DPR achieves better scalability than the exact algorithm and outperforms the approximate algorithm.
    \item We also introduce several baseline methods, including two RL-based heuristics, and evaluate them on multiple synthetic graphs and a real AD network. Our experimental results show that DPR consistently achieves superior performance over all other methods.
\end{itemize}


\section{Problem Formulation and Related Work}


\subsection{Problem Formulation}

The Adaptive Path Removal (APR) problem can be formally defined as follows: Given a directed attack graph $G = (V, E)$ with a source $s$ and a destination node $t$. Each edge $e \in E$ is associated with a confidence score, defined by a function $conf : E \mapsto [0, 1]$. Every round, the system will propose a simple path from the current attack graph. A simple path $p$ is defined as a sequence of edges $p = \langle (v_0 = s, v_1), (v_1, v_2), \ldots, (v_{k-1}, v_k = t) \rangle$ such that no edge is repeated, i.e., $((v_i, v_{i+1}) \neq (v_j, v_{j+1})), \forall ( i \neq j )$. When a path $p$ is queried, the IT admin selects exactly one edge $e \in p$ to remove. We model the IT admin’s choice using the Bradley–Terry model, which assigns a probability to each edge $e \in p$ proportional to its confidence score relative to the others in $p$: 

\begin{equation}
\label{eq:conf}
Pr(e|p) = \frac{conf(e)}{\sum_{e' \in p} conf(e')}
\end{equation}

In other words, an edge with a higher confidence score is more likely to be chosen for removal. This models the administrator’s relative preference or belief about which edge’s removal is most effective.
Let $C$ be the set of edges removed by the IT administrator after $|C|$ round. At round $|C| + 1$, the system will proposed a path $p \in P_{G'}$ in the temporary graph $G' = (V, E \setminus C)$ where $G'$ is called the temporary graph which evolved from the original graph $G = (V, E)$ by removing set of $C$ edges and $P_{G'}$ is the set of every possible path from $s$ to $t$ in $G'$. The query process terminates when either $C$ forms an $(s, t)$-cut (i.e., $s$ is disconnected from $t$) or the query budget $B$ is reached (i.e. $|C| = B$). In this paper, all cuts refer to $s-t$ cuts. To minimize human effort during the cutting process, our optimization goal is to design a policy that minimizes the expected number of queries (or iterations) required to complete the cutting process.


\begin{theorem}
    The APR Problem is $\#\mathcal{P}$-hard 
\end{theorem}
\begin{proof}
    We defer the proof the appendix Section \ref{sec:hardness}
\end{proof}



\textbf{Reason for edge's confidence score and how to assign it} Integrating confidence scores helps us effectively embed domain-specific security information into our model, making it easier to identify edges that are more likely to be removable. This matters because not all edges are equally prone to be removed by IT admin; some edges, such as outdated privileges or overly permissive group assignments, are clear candidates for removal. Using insights from the security context in our edge preference model could substantially reduce the number of required queries.
To automate the assignment of confidence scores, we can train a binary classifier that predicts the likelihood of each edge being safely removable. For example, Zheng et al.~\cite{zheng2011active}  propose an active learning approach that learns an IT admin’s decisions about which edges to remove. We can automatically assign confidence scores to edges by using a binary classifier, defined as a function $f : E \mapsto [0, 1]$ where the output represents the classifier’s confidence that a given edge can be safely removed.

\subsection{Related Works}

\textbf{Active Directory and non-adaptive defense models.}
The seminal work by Dunagan et al. \cite{dunagan2009heat} proposed the Active Directory (AD) attack graph which modelled the identity snowball attack that developed further and commercialized by Bloodhound \cite{BloodHound}. Follow-up works by Guo et al. \cite{guo2022practical,guo2023scalable} and Zhang et al. \cite{zhang2023oracle} formulate the problem of hardening the AD system as the shortest path interdiction via edge-removing problem. \cite{goel2022defending,goel2023evolving} proposed the Evolutionary Diversity Optimization (EDO) algorithm to defend against attackers in a configurable environment. Another work by Zhang et al. \cite{zhang2024practical} studied the problem of minimizing the number of users with paths to the domain admin via edge removal. Another approach for defending Active Directory found in the literature involves node-removal, which abstracts the concept of decoy allocation as introduced in Ngo et al. \cite{ngo2024catch,ngo2024optimizing}. The main drawback of non-adaptive models in real-world deployments is that they are not amenable to include human feedback.

\textbf{Adaptive models for Active Directory defense}. Several studies have integrated manual feedback from IT admin into network defenses process, emphasizing the importance of human involvement in configuration changes. 
Dunagan et al. \cite{dunagan2009heat} proposed Heat-ray, a system aimed at minimizing snowball identity attacks in Active Directory (AD) by iteratively proposing edge removals to IT administrators based on the sparest cut. 
Zheng et al. \cite{zheng2011active} enhanced Heat-ray with active learning to improve edge cost learning process. 
Guo et al. \cite{guo2024limited} introduced an adaptive defense model called the Limited Query Graph Connectivity Test (LQGCT), which is closely related to our approach. In their model, a proxy algorithm proposes one edge at a time, and the IT admin's decision is binary (i.e. whether to remove or retain it). By contrast, the proxy of our model proposes an entire attack path instead of a single edge which offers a multiple-choice selection rather than a binary decision.
Proposing a path provides several practical advantages over proposing an edge. Firstly, an edge proposal can fail to form a graph cut if the IT admin is overly conservative and retains too many edges.
This leaves the possibility of an attack even after the clean-up. In our experiments, path-based proposals guarantee that no attack path remains, provided the proxy algorithm has a sufficiently large budget.
Secondly, by presenting a list of edges to compare, our model encourages more deliberate choices, whereas a binary question as in LQGCT may incentivize conservative behaviour.
From a theoretical view, path-based proposals fundamentally differ and are harder to solve than the previous edge-based model. In LQGCT, the policy tree is binary, while our policy tree can branch into up to $l$ outcomes at each step, where $l$ is the length of the longest proposed path. As a result, existing algorithms cannot be directly applied to our setting, requiring us to develop an entirely new class of solutions.

\textbf{Related models from other research communities}. The sequential testing problem in operations research \cite{unluyurt2004sequential}, is often described through medical testing use cases. For instance, \cite{short2013iron,yu2023deep} employs adaptive strategies to reduce testing costs to diagnose diseases. Another related area is the problem of learning with attribute costs problem in machine learning \cite{sun1996hill,kaplan2005learning,golovin2011adaptive}. In this problem, each feature incurs a cost, and the task is to construct a classification tree that minimizes the total feature costs. Stochastic Boolean Function Evaluation (SBFE) problem \cite{allen2017evaluation,deshpande2014approximation} is also closely related. An SBFE instance involves a Boolean function $f$  with multiple hidden binary inputs and one binary output. Each input bit can be queried at a cost, and the objective is to find a query strategy that minimizes the expected cost to determine $f$'s output. 
While these models are relevant, they are not designed for our graph-based problems and lack scalability for large graphs. Consequently, similar to LQGCT, solutions for these models cannot be directly applied to our work.

\section{Algorithms}
In this part, we will present our solution for the APR problem. 
To help with the solution formulation, we will convert our problem into an equivalent Markov Decision Process (MDP). 
\subsection{Markov Decision Process formulation and preliminary}

Let us define the MDP as a tuple $\langle \mathcal{S}, \mathcal{A}, \Phi, R \rangle$, where $\mathcal{S}$ is the set of state, $\mathcal{A}$ is the set of action, $\Phi : \mathcal{S} \times \mathcal{A} \times \mathcal{S} \mapsto [0, 1 ]$ is the state transition and the reward function $R : \mathcal{S} \times \mathcal{A} \mapsto \mathbb{R}$. 

\textbf{State:} In an APR problem, the IT admin will remove an edge from a proposed path in every round. This process will evolve the graph into a series of temporary graphs by removing edges. We present these temporary graphs using a temporary state variable $s$ with $|B|$-dimension: $s = \{(x_1, x_2, \cdots, x_{B}) : x_i \in E \cup \{*\}, \forall i \in \{1, 2, \cdots, B\}\}$ where $x_i$ will be the edge that is removed by the IT admin at round $i$ and  $x_i = $ '*' means we have not query any path in this round.  We define the state of the original attack graph $G$ as the root state $s_r$, which will have the form: $(*, *, \cdots, *)$. A state $s'$ evolves from a state $s$ by removing an edge $e$ expressed as $s' = s \setminus e$. We denote $\mathcal{S}_i$ the set of possible states in round $i$ of the process. Hence, the state space can be represented as $\mathcal{S} = \mathcal{S}_0 \cup \mathcal{S}_{1} \cup \mathcal{S}_{2} \cup \cdots \cup \mathcal{S}_{B}$. We also define two sets of terminal states: $\perp_b$ is the terminal state reached when the budget is exhausted without identifying a cut, and $\perp_d$ is the terminal state reached when a cut is successfully found as a result of the query sequence. 

\textbf{Action:} Each path proposal in the APR problem is associated with an action in MDP. Let's say we are at state $s'$ associated with a temporary graph $G'$ and $A_{s'}$ is the action space at state $s'$. The action $a \in A_{s'}$ associates with a simple path $p \in P_{G'}$. We have the following Lemma for the action space in our problem:

\begin{lemma}
\label{lemma:subpath}
Given an MDP construction $\langle \mathcal{S}, \mathcal{A}, \Phi, R \rangle$ for the APR problem. We have $A_s \subseteq A_{s_r}$ for every $s \in \mathcal{S}$ where $s_r$ is the root state.
\end{lemma}

\begin{proof}
    The action available at each temporary state $s$ is the enumeration of every possible path in the corresponding graph $G$. A temporary graph $G'$ is actually the subgraph of the root graph $G$ (as $G'$ is evolved from $G$ by removing edge) which implies $P_{G'} \subseteq P_{G}$. Therefore, we have $A_{s} \subseteq A_{s'}$, $\forall s, s'$ where $s \in successor(s')$ which imply $A_s \subseteq A_{s_r}$. 
\end{proof}

Lemma \ref{lemma:subpath} states that action set $A_s$ of every state $s \in \mathcal{S}$ is a subset of the action set $A_{s_r}$ of the root state $s_r$. As a result, the overall action space for the APR problem can be expressed as $A = \bigcup_{s\in \mathcal{S}} A_s = A_{s_r}$. Lemma \ref{lemma:subpath} is particularly useful in the design of our algorithms, as discussed in the following sections. 

\textbf{Transition Probabilities:} In each state, an action can lead to different outcomes, which are defined by the transition probabilities in the MDP. In our problem, a transition probability shows the probability of an edge being removed by the IT admin when a path is proposed. The removal probability is defined by the Bradley-Terry preference model, as defined in Equation~\eqref{eq:conf}. Let's say we have a state $s' = s \setminus e$ where $s$ evolved to $s'$ by removing edge e. The transition probability from $s$ to $s'$ when taking action $a$ can be expressed as  $\Phi(s'|s,a) = \Phi(e|s, a) = Pr(e|p)$.

\textbf{Reward:} In our problem, each query will be penalized by a cost of exactly one unit of budget with no discount factor. The reward is difference at two terminal states: when $\perp_b$ is reached, meaning that we ran-out of budget before identifying any cut, we will penalize it with a constant of $\alpha > 0$; 
\begin{equation}
\label{eq:reward}
  R(s, a)=\begin{cases}
    -\alpha, & \text{if } s \in \perp_{b}.\\
    0, & \text{if } s \in \perp_{d}.\\
    -1, & \text{otherwise}.
  \end{cases}
\end{equation}

\textbf{Realization:} While "realization" is not a standard notation in MDP literature, it is commonly used in the context of adaptive submodularity optimization \cite{golovin2011adaptive}. We introduce this concept here as it will help us in describing the algorithms. We define a function $\phi : P_{s_r} \mapsto E$  as the full realization. We can view function $\phi(p)$ as an oracle that returns the IT administrator's edge removal decision for a given path $p$. Additionally, we define the partial realization $\psi_{s}: P_{s_r} \mapsto E$ as the observations made so far at state $s$. Specifically, $\psi_{s}(p)$ returns the edge removed by the IT administrator when $p$ is proposed, and $\psi_s(p) = *$ if $p$ has not yet been proposed or contains any edge have been removed by the IT admin. The domain of the partial realization is defined as  $dom(\psi) = \{p \in P_{s_r}: \psi(p) \neq *\}$, representing the set of actions for which outcomes have been observed. The range of the partial realization is defined as $range(\psi) = \{\psi(p) : p \in P_{s_r}, \psi(p) \neq *\}$, representing the edges that have been removed by IT admin. A partial realization $\psi$ is consistent with realization $\phi$ if they are equal everywhere in the $dom(\psi)$, denoted $\phi \sim \psi$. We say a partial realization $\psi$ is a subrealization of $\psi'$, denoted by $\psi \subseteq \psi'$, if they are both consistent with some $\phi$ and $dom(\psi) \subseteq dom(\psi')$.

\subsection{Dynamic Programming Exact Algorithm (OPT)}
\label{sec:opt}

As we establish the equivalence between the APR problem and the MDP, we also find that our problem satisfies the ``Principle of Optimality" in MDP. In our problem, given a state $s$ and $s' = s \setminus e$, $\forall e \in E_s$ and $E_s$ is the set of edges in the graph corresponding to state $s$, the optimal path query in $s$ is independent of previous queries and solving for the optimal strategy at $s'$ can be viewed as a subproblem of $s$. This allows us to introduce the optimal utility function following the Bellman equation:
\begin{equation}
\label{eq:dp}
    u_{\pi}(s) = \min_{p \in A_{\textbf{s}}}\{r(s, a) + \sum_{e \in p} \Phi (e|s,p) u_{\pi}(s\setminus e) \}, \text{  } \forall s \in \mathcal{S}
\end{equation}

Based on the the optimal utility function in \eqref{eq:dp}, we can design a Bellman's style dynamic programming, called OPT. Now, let $G_s$ be the graph that is associated with state $s$. The Dynamic Programming follows a top-down approach. In each subproblem, we require the utility $u_{\pi}(s)$. However, as shown in Equation \eqref{eq:dp}, obtaining the optimal utility requires invoking the action set $A_s$ for every subproblem, which in turn requires enumerating every simple path $P_{G_{s}}$ in $G_s$. This approach is impractical to run on any graph of reasonable size because the number of subproblems can grow as large as $|E|^B$, and path enumeration, known to be $\#\mathcal{P}$-hard, takes $\mathcal{O}(|V|^k)$ time complexity under a DFS-based approach \cite{peng2019hop}, where $k$ is the longest path length. However, thanks to Lemma \ref{lemma:subpath}, we can run the enumeration one time only for the original problem. The action space for the subproblem is simply $A_s = A_{s_r} \setminus \{p : e \in p, p \in A_{s_r}, e \in range(\psi_s)\}$, i.e., we can obtain $A_s$ by removing paths in $A_{s_r}$ that contains edges that have been chosen to be removed by the IT admin. Since running Depth First Search procedure to check if $s, t$ connectedness in every subproblem will take $\mathcal{O}(|E|+|V|)$. The overall Exact algorithm will take us $\mathcal{O}(|V|^{k}+(|E|+|V|)*|E|^B)$. We will defer the detailed pseudoscope of this algorithm to the appendix.

\subsection{Adaptive Submodular Approximation Algorithms (APP)}
\label{sec:appx}

In this section, we present an approximation algorithm, called APP, by utilizing the adaptive submodularity framework \cite{golovin2011adaptive}. The proposed algorithm provides square-logarithmic approximation to the number of enumerations of possible simple paths in the original graph $G$. The Adaptive Submodular Algorithm is proposed for the Stochastic Submodular Set Coverage (SSSC) problem in \cite{golovin2011adaptive} which has a close connection with APR problem.

\begin{Problem}The SSSC problem involves a ground set of elements $U = \{u_1, u_2, \cdots u_n\}$ and a collection of items $E = \{e_1, e_2, \cdots, e_m\}$, where each item $e$ is associated with a distribution over subsets of U. When an item is selected, a set is sampled from its distribution, i.e., it will reveal which subset of $U$ will be covered. The objective of this problem is to find an adaptive policy $\pi$ that selects items to cover all elements in $U$ while minimizing the expected number of items. To define the coverage, we define a utility function $f: 2^E \mapsto \mathbb{R}$ that quantifies the coverage achieved by the current state. The complete coverage is represented as states with utility meeting a predefined quota $Q$, i.e., $f(s) = Q$. 
\end{Problem}

We can see that our problem can be viewed as a special case of the SSSC problem. In the APR problem, the ground set consists of all simple paths $P_{s_r}$. In every round, when a path $p$ is proposed, the IT admin will choose an edge $e \in p$ to remove, every path in the set $P' = \{p' | e \in p', \forall p' \in P_{s_r} \}$ (the set of paths containing $e$) will also be removed. Each path in the action space can be viewed as an item in the SSSC problem, with each path associated with a distribution over the potential removal of other paths. This distribution is presented as in Equation \ref{eq:conf}. The goal of APR problem is to cover all paths in $P_{s_r}$ (a cut eliminates all paths) while minimizing the number of queries. Next, we have the following definitions.


\begin{definition}
    \textbf{(Conditional expected marginal benefit)} Given a state $s$, an action $a$ and a utility function $g$, the expected marginal benefit of $a$ is defined as:
    \begin{equation}
    \label{eq:marginal}
    \Delta (a\mid s) = \sum_{e \in a}  \{ \Phi (e\mid s,a) * \left[g(s\setminus e) - g(s) \right] \}
    \end{equation}
\end{definition}

\begin{definition}
\label{def:monotone}
    \textbf{(Adaptive Monotonicity)} A utility function $g : S \mapsto \mathbb{R}_{\geq 0}$ is adaptive monotone if the \textit{benefit of selecting an action is always nonnegative}. Formally, function $g$ is adaptive monotonic if $\forall s\in S$ and $\forall e \in \{e \mid e \in a, a \in A_s \}$, we have: 
    \begin{equation}
    \label{eq:monotone}
    g(s\setminus e) - g(s) \geq 0
    \end{equation}
\end{definition}

\begin{definition}
\label{def:submodular}
    \textbf{(Adaptive Submodular)} A utility function $g : S \mapsto \mathbb{R}_{\geq 0}$ is adaptive submodular if the marginal benefit of selecting an action does not increase as more actions are selected. Formally, function $g$ is adaptive submodular if for all temporary state $s$, $s'$ such that $\psi_{s} \subseteq \psi_{s'}$, $\forall a \in A_{s_r}$, we have:
    \begin{equation}
    \label{eq:submodular}
    \Delta(a\mid s) \geq \Delta(a \mid s')
    \end{equation}
\end{definition}

\textbf{The reason for introducing these concepts} is to port algorithms from the Adaptive Submodular framework to the APR problem while ensuring the theoretical approximation bound. To do so, we need to design a utility function $g$ associated with the APR problem that satisfies two key conditions: (1) adaptive monotonicity and (2) adaptive submodularity.

\textbf{Utility Function:} The utility function $g : S \mapsto \mathbb{N}$ is defined as:
    \begin{equation}
    \label{eq:utility}
        g(s) = |\bigcup_{e \in range(\psi_{s})} h(e, P_{s_r})|
    \end{equation}

where the function  $h(e, P) = \{p | e \in p, \forall p \in P\}$ returns the set of paths $p\in P$ that contains $e$. To remind, $range(\psi_{s})$ is the set of edges that have been removed upon state $s$. We have the following lemma for the utility function $g$:

\begin{lemma}
\label{lemma:g}
    Function $g$ is both adaptive monotonic and adaptive submodular 
\end{lemma}
   \begin{proof}

        First, consider the function $g$, which counts the number of paths removed up to the current state. Suppose we are at any state $s$, and path $p$  is proposed to the IT admin, who then chooses to remove an edge $e$. Removing $e$ eliminates at least the proposed path $p$ since $e \in p$. This means $g(s \setminus e) \geq  g(s) + 1$ and satisfying Equation \eqref{eq:monotone}. Therefore, $g$ is adaptive monotone. 

        Moreover, since the utility function $g$, as defined in Eq. \eqref{eq:utility}, returns the number of paths removed by IT admin decision from the root state. Furthermore, removing an edge $e$ in a state $s$ with $\psi_s \subseteq \psi_{s'}$ will result in more paths being eliminated than removing $e$ in the successor state $s'$ which is formally expressed as $g(s\setminus e) - g(s) \geq g(s'\setminus e) - g(s')$. This implies: 

        \begin{equation}
        \nonumber
        \begin{split}
            \Delta(a \mid s) &=  \sum_{e \in a}  \{ \Phi (e\mid s,a) * \left[g(s\setminus e) - g(s) \right] \} \\
                            &\geq \sum_{e \in a}  \{ \Phi (e\mid s',a) * \left[g(s'\setminus e) - g(s') \right]\} = \Delta(a \mid s')
        \end{split}
        \end{equation}
        The transition from line 1 to line 2 is valid because $\Phi (e\mid s,a) = \Phi (e\mid s',a)$, $\forall s, s', a$ as the transition probabilities of an action do not depend on the state, as defined in the equation \ref{eq:conf}. This proves the adaptive submodularity of $g$.
\end{proof}

Note, to ensure Theorem \ref{theorem:approx1} holds, $g$ must be both \textit{strongly adaptive monotonic and submodular}. While Definition \ref{def:monotone} aligns with the concept of strongly adaptive monotonicity \cite{golovin2011adaptive}, Definition \ref{def:submodular} is only adaptive submodularity. A function is strongly adaptive submodular if it is (1) adaptive submodular and (2) pointwise submodular. Although we have proven the former, we admit the second property for $g$ and hence $g$ is also strongly adaptive submodular.
We defer the definition and proof of (2) to the appendix to maintain the readability of the paper.

\textbf{Greedy with marginal gain strategy (Algorithm \ref{alg:greedy}):} By applying a greedy strategy with our problem-tailored utility function $g$, we have our Adaptive Submodular Algorithm as shown in Algorithm \ref{alg:greedy}.
In this algorithm, during each query round, we greedily propose the action $a \in A_{s}$ that yields the highest expected marginal benefit with respect to the utility function $g$. 
Here again, due to Lemma \ref{lemma:subpath}, we only need to enumerate the action space (for the $G_{s_r}$) once beforehand.
Note, while the path enumeration problem is known to be $\#\mathcal{P}$-hard, our experiments with attack graphs demonstrate that this enumeration can be performed within a reasonable runtime. The efficiency of this process is largely because attack graphs typically involve only subgraphs of the larger AD structure. 

\begin{algorithm}[H]
 \caption{Adaptive Submodular Strategy (APP)}
 \label{alg:greedy}
 \begin{algorithmic}[1]
 \renewcommand{\algorithmicrequire}{\textbf{Input:}}
 \REQUIRE Directed graph $G(V, E)$, source $s$ and destination $t$ \\
 \renewcommand{\algorithmicrequire}{\textbf{Output:}}
 \REQUIRE approximate propose strategy \\
 \STATE Initialise set of simple path $A_{s_r}$ of original state $s_r$, set of proposed path $A_{\pi}$
 \STATE \textbf{while} $s$ is not terminate state
 \STATE \quad \textbf{foreach} $a \in A_{s_r}\setminus A_{\pi}$
 \STATE \quad \quad $\Delta(a|s) = \sum_{e \in a}  \{ \Phi (e|s,a) * \left[g(s\setminus e) - g(s) \right] \}$
 \STATE \quad $a^{*} = \arg\max_a \Delta(a|s)$
 \STATE \quad $A_{\pi} = A_{\pi} \cup \{a^*\}$, proposed $a^*$, observe outcome $e*$
 \STATE \quad $s = s \setminus e^*$, progressing to new state
 \end{algorithmic}
\end{algorithm}

Once we have proven that $g$ is both adaptive monotonic and adaptive submodular in Lemma \ref{lemma:g}, the following theoretical approximation ratio for Algorithm \ref{alg:greedy} follows.

\begin{theorem}
\label{theorem:approx1}
 Algorithm \ref{alg:greedy} achieves a $(\ln{|P_{s_r}| + 1)^2}$-approximation for the APR problem with $B = |P_{s_r}|$.    
\end{theorem}

\subsection{Scalable Heuristics Algorithm (DPR)}

In this section, we present a scalable heuristic designed based on the exact algorithm and the approximate algorithm as shown in Algorithm \ref{alg:dpheu}


\textbf{Heuristics based on Exact Algorithm (DPR)} The exact dynamic programming algorithm struggles to scale in realistic scenarios due to two main issues. First, APR problem's MDP often has large action space, particularly in the initial states, where the action space size corresponds to the enumeration of simple paths in the original graph. Second, for each action, the number of child subproblems to solve can grow up to $\mathcal{O}(|E|^k)$. We proposed a scalable heuristic in Algorithm \ref{alg:dpheu}, designed to restrict the subproblem space in dynamic programming to a manageable size and enable efficient execution on graphs of practical scale. We call this algorithm dynamic programming with restriction (DPR). Function $DPR(s', r, B')$ in Algorithm \ref{alg:dpheu} shows a modification of the Exact Dynamic Programming algorithm with restriction. The first restriction is that instead of considering subproblems from $B$ steps ahead, \textit{DPR} reduces the lookahead to $B'$ steps, where $B' < B$. The second restriction is to avoid enumerating every possible state (which can be problematic in the early stages). Instead, we only consider a set of $\tau$ candidate paths, implemented as the $path\_sampling(A_{s'}, \tau)$ function in Algorithm \ref{alg:dpheu}. In general, we modify each heuristic to return the top $k <\tau$ candidate paths, rather than a single best path, based on each heuristic’s ranking criterion. For example, the approximate heuristic ranks paths by their marginal gain according to $g$, then selects the top $k$ among them. This function will draw from multiple heuristic methods—such as those derived from an approximation strategy (Algorithm \ref{alg:greedy}), shortest paths, approximate strategies on sets of shortest paths, or paths likely to remove an edge in a minimum cut. As our experiments show that these approaches provide strong performance. This modification decreases the number of subproblems to $\mathcal{O}((\tau)^{B'})$ for each DP, making DP more feasible in larger settings.
While this adjustment may affect the optimality of the solution (compared to the exact DP algorithm), it significantly improves the scalability. Empirically, we will experimentally demonstrate that the \textit{DPR} algorithm scales better than the exact algorithm and outperforms the approximate algorithm on every graph.

\begin{algorithm}[H]
 \caption{Heuristics based on Exact Algorithm (DPR)}
 \label{alg:dpheu}
 \begin{algorithmic}[1]
 \renewcommand{\algorithmicrequire}{\textbf{Input:}}
 \REQUIRE Directed Attack Graph $G(V, E)$, budget $B$, look-ahead budget $B'$
 \renewcommand{\algorithmicrequire}{\textbf{Output:}}
 \REQUIRE Heuristic query strategy \\
 \STATE \textbf{while} $\textbf{s}$ is not a terminate state
 \STATE \quad $\pi = DPR (\textbf{s}, |A_{\pi}|, B')$
 \STATE \quad $a^* = \pi(\textbf{s})$
 \STATE \quad $A_{\pi} = A_{\pi} \cup \{a^*\}$, proposed $a^*$, observe outcome $e*$
 \STATE \quad $\textbf{s} = \textbf{s} \setminus e^*$, progressing to new state
 \STATE 
 \STATE \textbf{function} $DPR (\textbf{s}', r, B', \tau)$
 \STATE \quad \textbf{for} $i \in \left[ 0, B'\right]$
 \STATE \quad \quad \textbf{for} $\textbf{s}' \in \mathcal{S}_{|A_{\pi}| + i}$
 \STATE \quad \quad \quad \textbf{if} $\textbf{s}'$ is in $\perp_{b}$ or $\perp_{d}$
 \STATE \quad \quad \quad \quad $u_{\pi}(\textbf{s}'\setminus e)=\begin{cases} 
    \alpha, & \text{if } \textbf{s}' \in \perp_{b}\\
    0, & \text{if } \textbf{s}' \in \perp_{d}\\
    \end{cases}$ \\
 \STATE \quad \quad \quad \quad $\pi(\textbf{s}'\setminus e) = \emptyset$
 \STATE \quad \quad \quad \textbf{else}
 \STATE \quad \quad \quad \quad $A = path\_sampling(A_{\textbf{s}'}, \tau)$
 \STATE \quad \quad \quad \quad $a^* = \operatorname*{argmin}_{p \in A}\{ \sum_{e \in p}\left[  \Phi(e|\textbf{s}',a) * u_{\pi}(\textbf{s}'\setminus e)\right]\}$
  \STATE \quad \quad \quad \quad $u_{\pi}(\textbf{s}') = 1 + \sum_{e \in p}\left[   \Phi (e|\textbf{s}',a) * u_{\pi}(\textbf{s}'\setminus e)\right]$
 \STATE \quad \quad \quad \quad $\pi(\textbf{s}') = p^*$
 \STATE \quad \textbf{return} $\pi$
 \end{algorithmic}
\end{algorithm}

\section{Experiment}

In this section, we present the evaluation of our algorithm on 13 synthetic graphs of different sizes and an Active Directory (AD) attack graph from a real organization. 
\subsection{Experiment Set Up}

In our experiment, we evaluate our algorithm using synthetic AD attack graphs generated by ADSynth \cite{nguyen2024adsynth}, a state-of-the-art AD graph generator. ADSynth models AD graphs based on Microsoft’s best practices tiering model \cite{bestpractice,bestpractice2}, where Tier 0 contains the highest privilege nodes with administrative control, Tier 1 includes high-privileged servers, and Tier 2 and beyond contain non-administrative nodes. ADSynth simulates the AD attack graph in two steps: (1) generating a best-practice AD infrastructure and (2) creating cross-tier edges. 

If every node had a predefined tier, the defense problem would become trivial, as attack paths could be easily identified and removed by removing all edges connecting lower-privilege nodes to higher-privilege nodes \cite{improhound}. 
Open-source tools like ImproHound \cite{improhound} are designed to automate this process. 
However, assigning roles to nodes is inherently challenging due to the dynamic nature of roles, overlapping responsibilities, and exceptions such as temporary access \cite{tiertalk}. We called these \textit{undefined tier nodes}. 
In our simulated attack graph, we assume the presence of a set of nodes with undefined tier connections which create attack paths from lower-privilege nodes to higher-privilege nodes. We assume that IT admin use our adaptive model with the goal of removing all attack paths from the lowest tier to Tier 0. 
Our model treats the attack graph as a single-source, single-target graph so we merge all Tier 0 nodes into a single supernode $t$ and all lowest-tier nodes into a single supernode $s$. 

For our synthetic attack graphs, we labelled them from $G1$ to $G9$. In these graphs, the number of tiers is fixed at 3, and $95\%$ of nodes in graph have well-defined tier assignments. Additionally, we also have 4 smaller versions of the graph denoted from $GS1$ to $GS4$, used in the experiment in Table \ref{tab:smallgraph}. In this small graph, defined-role ratio is about $99\%$. We also included one real AD graph that we collected from an anonymous organization, we denoted this graph as ORG. 

All of the experiments are carried out on a high-performance computing cluster with 1 CPU and 24GB of RAM allocated to each trial. In Tables \ref{tab:smallgraph} and \ref{tab:biggraph}, we report the average number of queries over 16,000 trials. The budget constraint $B$ is set at 10 for all experiments on synthetic graphs. For the real AD graph ORG, due to computing resource limitations, we report the average number of queries over 200 trials. Also for ORG, we reserve a higher budget of 20 and 30 queries due to the size of this graph, denoted ORG(20) and ORG(30) respectively. For the DPR algorithm, we set $\tau = 16$ actions and a lookahead budget of $B' = 4$ step. We reserve a higher budget of 20 and 30 queries due to the size of this graph. 
For the DPR algorithm, we set $\tau = 16$ actions and a lookahead budget of $B' = 4$ step. 

\begin{table}[h]
\begin{center}
\caption{Expected number of query under different algorithm \textbf{($\downarrow$ is better)}. Here, we only consider graphs where OPT can run on. }\label{tab:smallgraph}
\smallskip\noindent
\resizebox{0.35\linewidth}{!}{%
\begin{tabular}{ccccccc}
\toprule
        ~      & \textbf{$GS1$} & \textbf{$GS2$} & \textbf{$GS3$} & \textbf{$GS4$}  \\ \hline
        OPT    & \textbf{2.513}  & \textbf{2.592}  & \textbf{2.545}  & \textbf{2.383}  \\ \hline
        APP& \textbf{2.513}  & \textbf{2.592}  & 2.546           & 2.385  \\ \hline
        OTH1 & \textbf{2.513}  & \textbf{2.592}  & 2.546          & \textbf{2.383}  \\ \hline
        OTH2 & \textbf{2.513}  & 2.596           & \textbf{2.545} & 2.388  \\ \hline
        PPO    & \textbf{2.513}  & \textbf{2.592}  & 2.546           & \textbf{2.383}  \\ \hline
        SAC    & 2.514           & \textbf{2.592}  & 2.546           & 2.384  \\ \hline
        DPR   & \textbf{2.513}  & \textbf{2.592}  & 2.546           & \textbf{2.383} \\ 
\toprule
\end{tabular}}
\end{center}
\end{table} 

\begin{table*}[h]
\begin{center}
\caption{Expected number of query under different algorithm \textbf{($\downarrow$ is better)}. AVG.RANK represents the average head-to-head performance ranking of each algorithm across all evaluated graphs. $\#n/\#e$ show the number of nodes and edge in the graph. MC is the min-cut. }\label{tab:biggraph}
\smallskip\noindent
\resizebox{1\linewidth}{!}{%
\begin{tabular}{cccccccccccc|c}
\toprule
        ~      & \textbf{G1}& \textbf{G2} & \textbf{G3} & \textbf{G4} & \textbf{G5} & \textbf{G6} & \textbf{G7} & \textbf{G8} & \textbf{G9} & \textbf{ORG}(20)& \textbf{ORG}(30) & AVG.RANK  \\ 
        \#n/\#e   & 1047/5078              & 1047/5091               & 1047/5116               & 5147/25376             & 5139/25153              & 5139/25161              & 10070/48161               & 10070/48170               & 10070/48192              & 125444/1195432              & \_              &  \_     \\ 
        MC   & 3               & 3                & 3                & 3                & 3                & 4
               & 3               & 3                & 3                & 8                & \_  & \_ \\ \hline

        APP& 3.821           & 3.762            & 4.534            & 4.334            & 3.879            & 4.594            & 3.807             & 3.869             & 3.590            & 17.605           & 18.840           & 3.889   \\ \hline
        OTH1 & 3.816           & \textbf{3.755}   & \textbf{4.409}   & \textbf{3.885}   & 3.880            & 4.593            & 3.810             & 3.893             & 3.584            & 17.485           & 18.600           & 3.185  \\ \hline
        OTH2 & \textbf{3.813}  & 3.756            & 4.437            & 3.904            & 3.883            & 4.592            & 3.799             & 3.871             & 3.570            & 17.535           & \textbf{18.535 }          & 3.333  \\ \hline
        PPO    & 3.816           & \textbf{3.755}   & 4.425            & 3.905            & 3.876            & 4.587            & 3.797             & 3.876             & 3.573            & 17.665           & 18.835           & 2.667  \\ \hline  
        SAC    & 3.854           & 3.799            & 4.490            & 3.901            & \textbf{3.874}   & 4.606            & \textbf{3.792}           & 3.876            & 3.575            &  18.005          & 18.560            & 2.667  \\ \hline
        DPR   & 3.816           & \textbf{3.755}   & \textbf{4.409}   & 3.901            & 3.876            & \textbf{4.589}   & 3.797             & \textbf{3.869}    & \textbf{3.568}   &  \textbf{17.480}          & 18.555          & \textbf{1.444}  \\ 
\toprule
\end{tabular}}
\end{center}

\end{table*}

\subsection{Baseline Algorithms}

\textbf{Reinforcement Learning.} This approach shares a similar concept with DPR but replaces the use of Dynamic Programming with restricted lookahead by a model-free reinforcement learning to learn the query strategy. We utilize two model-free reinforcement learning models: Proximal Policy Optimization (PPO) \cite{schulman2017proximal} and Soft Actor-Critic for Discrete Action (SAC) \cite{christodoulou2019soft}. We encode the observation space as a vector of $(E + \tau B)$ binary bits. The first $E$ bits represent a one-hot encoding of the edges that have been removed through queries, while the remaining $\tau B$ bits encode the taken actions. We allocated a100 GPUs for the training of RL agents.

\textbf{Others Heuristics} We also introduce two other heuristics called OTH1 and OTH2 which are designed based on the approximate algorithm. For OTH1, we modify the utility function to ensure it will propose paths with the highest likelihood of removing an edge in the minimum cut set. Formally, it selects the path $a = \arg\max_{a \in P_{G'}:a \cap mc(G')} \sum_{e \in a}  \{ \Phi (e|s,a) * \left[g(s\setminus e) - g(s) \right] \}$ where $mc(G')$ return the $s-t$ minimum cut of the temporary graph $G'$. For the OTH2, we restrict the approximate algorithm to run on the set of shortest paths only. 

\subsection{Performance Interpretation}

In Table \ref{tab:smallgraph}, we report the performance of our proposed algorithm in the graph where OPT can optimally come up with the query policy without out-of-memory error. As we mentioned, OPT is very costly computationally, we are only able to scale it to a graph with 17 nodes, 32 edges and 16 attack paths ($GS4$ graph). Overall, all of our heuristics (OTHs, PPO, SAC and DPR) perform very well with the small optimality gap.

In Table \ref{tab:biggraph}, we present the performance of our algorithm on 13 large synthetic graphs and one real-world AD graph (\textbf{ORG}) from an anonymous organization. We observed that DPR consistently achieved the best performance. Noticeably, DPR outperformed APP in every graph. All proposed algorithms outperformed APP. Nevertheless, APP's performance is theoretically guaranteed which may be useful for some worst-case scenarios. This algorithm is also useful as a path sampling scheme for the DPR algorithm as the action space of DPR algorithm contain the approximate strategy which somewhat helps DPR to have a guaranteed performance. While the RL algorithms (PPO and SAC) performed well on synthetic graphs, their performance was worse on the real AD graph. We suspect this is due to the real graph requiring a significantly larger number of queries. The RL policies are myopic, meaning they excel in scenarios with fewer queries by prioritizing short-term gains but struggle when a higher number of queries is needed, as they fail to account for long-term gains.   

The adaptive hardening of AD security has been studied as the LQGCT problem by Guo et al. \cite{guo2024limited}. Their model simulates IT admin' behaviour as a binary decision-making process: at each step, an edge is queried, and the administrator labels it as either "cut" or "retain." However, this approach often fails when IT admins are overly conservative, retaining too many edges and leading to unsuccessful cuts. In contrast, our model queries a path in each step, presenting a multiple-choice decision for the IT admin to select one edge to remove. In table \ref{tab:guocomp}, we compare the graph-cutting performance of the RL algorithm from LQGCT with our DPR algorithm. The results from 512 trials demonstrate that our model achieved more successful cuts compared to Guo's model. We observe that a larger budget leads to a higher cutting success rate in our model, this means we can guarantee a successful cut in every trial by allocating a sufficiently larger budget.

\begin{table}[h]
\begin{center}
\caption{The number of trials with successful cuts between RL from LQGCT model and DPR algorithm over 512 trials. \textbf{($\uparrow$ is better)}}\label{tab:guocomp}
\smallskip\noindent
\small
\resizebox{0.4\linewidth}{!}{%
\begin{tabular}{ccccc}
\toprule
       &\multicolumn{2}{c}{LQGCT's RL }&\multicolumn{2}{c}{DPR}\\
                        & B = 5 & B = 10 & B = 5 & B = 10 \\
        \hline
       \textbf{G7}&  128  & 144    &   278 &  459   \\ \hline
       \textbf{G8}&  96   &  96    &   419 &  504   \\ \hline
       \textbf{G9}&  96   &  96    &   339 &  487   \\ 
\toprule
\end{tabular}}
\end{center}
\end{table}


\section{Conclusion}

In this paper, we proposed a practical human-in-the-loop combinatorial problem for network security called Adaptive Path Removal problem. This problem was motivated by the technical requirements and limitations of current industrial models. The goal of our model is to reduce the workload for security teams in an adaptive manner. 
We proposed a comprehensive set of solutions, including an exact algorithm, an approximate algorithm, and several scalable heuristics.
Among these, our DPR heuristic, designed based on both the exact and approximate algorithms, exhibited superior performance.
Specifically, DPR demonstrated the ability to run effectively on larger-scale graphs compared to the exact algorithm and consistently outperformed the approximate algorithm across all tested graph scenarios.
We verify the effectiveness of our algorithm on several synthetic AD graphs and an AD attack graph collected from a real organization.

\bibliographystyle{ACM-Reference-Format}
\bibliography{main}


\begin{thebibliography}{30}


\ifx \showCODEN    \undefined \def \showCODEN     #1{\unskip}     \fi
\ifx \showDOI      \undefined \def \showDOI       #1{#1}\fi
\ifx \showISBNx    \undefined \def \showISBNx     #1{\unskip}     \fi
\ifx \showISBNxiii \undefined \def \showISBNxiii  #1{\unskip}     \fi
\ifx \showISSN     \undefined \def \showISSN      #1{\unskip}     \fi
\ifx \showLCCN     \undefined \def \showLCCN      #1{\unskip}     \fi
\ifx \shownote     \undefined \def \shownote      #1{#1}          \fi
\ifx \showarticletitle \undefined \def \showarticletitle #1{#1}   \fi
\ifx \showURL      \undefined \def \showURL       {\relax}        \fi
\providecommand\bibfield[2]{#2}
\providecommand\bibinfo[2]{#2}
\providecommand\natexlab[1]{#1}
\providecommand\showeprint[2][]{arXiv:#2}

\bibitem[Allen et~al\mbox{.}(2017)]%
        {allen2017evaluation}
\bibfield{author}{\bibinfo{person}{Sarah~R Allen}, \bibinfo{person}{Lisa Hellerstein}, \bibinfo{person}{Devorah Kletenik}, {and} \bibinfo{person}{Tongu{\c{c}} {\"U}nl{\"u}yurt}.} \bibinfo{year}{2017}\natexlab{}.
\newblock \showarticletitle{Evaluation of monotone DNF formulas}.
\newblock \bibinfo{journal}{\emph{Algorithmica}}  \bibinfo{volume}{77} (\bibinfo{year}{2017}), \bibinfo{pages}{661--685}.
\newblock


\bibitem[Ball(1986)]%
        {reliability}
\bibfield{author}{\bibinfo{person}{Michael~O Ball}.} \bibinfo{year}{1986}\natexlab{}.
\newblock \showarticletitle{Computational complexity of network reliability analysis: An overview}.
\newblock \bibinfo{journal}{\emph{Ieee transactions on reliability}} \bibinfo{volume}{35}, \bibinfo{number}{3} (\bibinfo{year}{1986}), \bibinfo{pages}{230--239}.
\newblock


\bibitem[Christodoulou(2019)]%
        {christodoulou2019soft}
\bibfield{author}{\bibinfo{person}{Petros Christodoulou}.} \bibinfo{year}{2019}\natexlab{}.
\newblock \showarticletitle{Soft actor-critic for discrete action settings}.
\newblock \bibinfo{journal}{\emph{arXiv preprint arXiv:1910.07207}} (\bibinfo{year}{2019}).
\newblock


\bibitem[Deshpande et~al\mbox{.}(2014)]%
        {deshpande2014approximation}
\bibfield{author}{\bibinfo{person}{Amol Deshpande}, \bibinfo{person}{Lisa Hellerstein}, {and} \bibinfo{person}{Devorah Kletenik}.} \bibinfo{year}{2014}\natexlab{}.
\newblock \showarticletitle{Approximation algorithms for stochastic boolean function evaluation and stochastic submodular set cover}. In \bibinfo{booktitle}{\emph{Proceedings of the twenty-fifth annual ACM-SIAM Symposium on Discrete Algorithms}}. SIAM, \bibinfo{pages}{1453--1466}.
\newblock


\bibitem[Dunagan et~al\mbox{.}(2009)]%
        {dunagan2009heat}
\bibfield{author}{\bibinfo{person}{John Dunagan}, \bibinfo{person}{Alice~X Zheng}, {and} \bibinfo{person}{Daniel~R Simon}.} \bibinfo{year}{2009}\natexlab{}.
\newblock \showarticletitle{Heat-ray: combating identity snowball attacks using machinelearning, combinatorial optimization and attack graphs}. In \bibinfo{booktitle}{\emph{Proceedings of the ACM SIGOPS 22nd symposium on Operating systems principles}}. \bibinfo{pages}{305--320}.
\newblock


\bibitem[Goel et~al\mbox{.}(2023)]%
        {goel2023evolving}
\bibfield{author}{\bibinfo{person}{Diksha Goel}, \bibinfo{person}{Aneta Neumann}, \bibinfo{person}{Frank Neumann}, \bibinfo{person}{Hung Nguyen}, {and} \bibinfo{person}{Mingyu Guo}.} \bibinfo{year}{2023}\natexlab{}.
\newblock \showarticletitle{Evolving Reinforcement Learning Environment to Minimize Learner's Achievable Reward: An Application on Hardening Active Directory Systems}.
\newblock \bibinfo{journal}{\emph{GECCO '23: Genetic and Evolutionary Computation Conference, 2023, 2023}} (\bibinfo{year}{2023}).
\newblock


\bibitem[Goel et~al\mbox{.}(2022)]%
        {goel2022defending}
\bibfield{author}{\bibinfo{person}{Diksha Goel}, \bibinfo{person}{Max~Hector Ward-Graham}, \bibinfo{person}{Aneta Neumann}, \bibinfo{person}{Frank Neumann}, \bibinfo{person}{Hung Nguyen}, {and} \bibinfo{person}{Mingyu Guo}.} \bibinfo{year}{2022}\natexlab{}.
\newblock \showarticletitle{Defending active directory by combining neural network based dynamic program and evolutionary diversity optimisation}. In \bibinfo{booktitle}{\emph{Proceedings of the Genetic and Evolutionary Computation Conference}}. \bibinfo{pages}{1191--1199}.
\newblock


\bibitem[Golovin and Krause(2011)]%
        {golovin2011adaptive}
\bibfield{author}{\bibinfo{person}{Daniel Golovin} {and} \bibinfo{person}{Andreas Krause}.} \bibinfo{year}{2011}\natexlab{}.
\newblock \showarticletitle{Adaptive submodularity: Theory and applications in active learning and stochastic optimization}.
\newblock \bibinfo{journal}{\emph{Journal of Artificial Intelligence Research}}  \bibinfo{volume}{42} (\bibinfo{year}{2011}), \bibinfo{pages}{427--486}.
\newblock


\bibitem[Guo et~al\mbox{.}(2022)]%
        {guo2022practical}
\bibfield{author}{\bibinfo{person}{Mingyu Guo}, \bibinfo{person}{Jialiang Li}, \bibinfo{person}{Aneta Neumann}, \bibinfo{person}{Frank Neumann}, {and} \bibinfo{person}{Hung Nguyen}.} \bibinfo{year}{2022}\natexlab{}.
\newblock \showarticletitle{Practical fixed-parameter algorithms for defending active directory style attack graphs}. In \bibinfo{booktitle}{\emph{Proceedings of the AAAI Conference on Artificial Intelligence}}, Vol.~\bibinfo{volume}{36}. \bibinfo{pages}{9360--9367}.
\newblock


\bibitem[Guo et~al\mbox{.}(2024)]%
        {guo2024limited}
\bibfield{author}{\bibinfo{person}{Mingyu Guo}, \bibinfo{person}{Jialiang Li}, \bibinfo{person}{Aneta Neumann}, \bibinfo{person}{Frank Neumann}, {and} \bibinfo{person}{Hung Nguyen}.} \bibinfo{year}{2024}\natexlab{}.
\newblock \showarticletitle{Limited Query Graph Connectivity Test}. In \bibinfo{booktitle}{\emph{Proceedings of the AAAI Conference on Artificial Intelligence}}, Vol.~\bibinfo{volume}{38}. \bibinfo{pages}{20718--20725}.
\newblock


\bibitem[Guo et~al\mbox{.}(2023)]%
        {guo2023scalable}
\bibfield{author}{\bibinfo{person}{Mingyu Guo}, \bibinfo{person}{Max Ward}, \bibinfo{person}{Aneta Neumann}, \bibinfo{person}{Frank Neumann}, {and} \bibinfo{person}{Hung Nguyen}.} \bibinfo{year}{2023}\natexlab{}.
\newblock \showarticletitle{Scalable edge blocking algorithms for defending active directory style attack graphs}. In \bibinfo{booktitle}{\emph{Proceedings of the AAAI Conference on Artificial Intelligence}}, Vol.~\bibinfo{volume}{37}. \bibinfo{pages}{5649--5656}.
\newblock


\bibitem[Kaplan et~al\mbox{.}(2005)]%
        {kaplan2005learning}
\bibfield{author}{\bibinfo{person}{Haim Kaplan}, \bibinfo{person}{Eyal Kushilevitz}, {and} \bibinfo{person}{Yishay Mansour}.} \bibinfo{year}{2005}\natexlab{}.
\newblock \showarticletitle{Learning with attribute costs}. In \bibinfo{booktitle}{\emph{Proceedings of the thirty-seventh annual ACM symposium on Theory of computing}}. \bibinfo{pages}{356--365}.
\newblock


\bibitem[Knudsen(2021)]%
        {improhound}
\bibfield{author}{\bibinfo{person}{Jonas~Bülow Knudsen}.} \bibinfo{year}{2021}\natexlab{}.
\newblock \bibinfo{title}{ImproHound: Identify the attack paths in BloodHound breaking your AD tiering}.
\newblock \bibinfo{howpublished}{\url{https://github.com/improsec/ImproHound}}.
\newblock


\bibitem[Knudsen and Schmitt(2023)]%
        {tiertalk}
\bibfield{author}{\bibinfo{person}{Jonas~Bülow Knudsen} {and} \bibinfo{person}{Alexander Schmitt}.} \bibinfo{year}{2023}\natexlab{}.
\newblock \bibinfo{title}{Hidden Pathways: Exploring the Anatomy of ACL-Based Active Directory Attacks and Building Strong Defenses}.
\newblock \bibinfo{howpublished}{\url{https://troopers.de/troopers23/talks/33fcyz/}}.
\newblock


\bibitem[Microsoft(2024a)]%
        {bestpractice}
\bibfield{author}{\bibinfo{person}{Microsoft}.} \bibinfo{year}{2024}\natexlab{a}.
\newblock \bibinfo{title}{Best Practice Guide for Securing Active Directory Installations.}
\newblock \bibinfo{howpublished}{\url{https://learn.microsoft.com/en-us/windows-server/identity/identity-and-access}}.
\newblock


\bibitem[Microsoft(2024b)]%
        {bestpractice2}
\bibfield{author}{\bibinfo{person}{Microsoft}.} \bibinfo{year}{2024}\natexlab{b}.
\newblock \bibinfo{title}{Mitigating Pass-the-Hash (PtH) Attacks and Other Credential Theft, Version 1 and 2}.
\newblock \bibinfo{howpublished}{\url{https://www.microsoft.com/en-au/download/details.aspx?id=36036}}.
\newblock


\bibitem[Ngo et~al\mbox{.}(2024a)]%
        {ngo2024optimizing}
\bibfield{author}{\bibinfo{person}{Huy Ngo}, \bibinfo{person}{Mingyu Guo}, {and} \bibinfo{person}{Hung Nguyen}.} \bibinfo{year}{2024}\natexlab{a}.
\newblock \showarticletitle{Optimizing Cyber Response Time on Temporal Active Directory Networks Using Decoys}. In \bibinfo{booktitle}{\emph{Proceedings of the Genetic and Evolutionary Computation Conference}}. \bibinfo{pages}{1309--1317}.
\newblock


\bibitem[Ngo et~al\mbox{.}(2024b)]%
        {ngo2024catch}
\bibfield{author}{\bibinfo{person}{Huy~Q Ngo}, \bibinfo{person}{Mingyu Guo}, {and} \bibinfo{person}{Hung Nguyen}.} \bibinfo{year}{2024}\natexlab{b}.
\newblock \showarticletitle{Catch Me if You Can: Effective Honeypot Placement in Dynamic AD Attack Graphs}. In \bibinfo{booktitle}{\emph{IEEE INFOCOM 2024-IEEE Conference on Computer Communications}}. IEEE, \bibinfo{pages}{451--460}.
\newblock


\bibitem[Nguyen et~al\mbox{.}(2024)]%
        {nguyen2024adsynth}
\bibfield{author}{\bibinfo{person}{Nhu~Long Nguyen}, \bibinfo{person}{Nickolas Falkner}, {and} \bibinfo{person}{Hung Nguyen}.} \bibinfo{year}{2024}\natexlab{}.
\newblock \showarticletitle{ADSynth: Synthesizing Realistic Active Directory Attack Graphs}.
\newblock  (\bibinfo{year}{2024}), \bibinfo{pages}{66--74}.
\newblock


\bibitem[Peng et~al\mbox{.}(2019)]%
        {peng2019hop}
\bibfield{author}{\bibinfo{person}{You Peng}, \bibinfo{person}{Ying Zhang}, \bibinfo{person}{Xuemin Lin}, \bibinfo{person}{Wenjie Zhang}, \bibinfo{person}{Lu Qin}, {and} \bibinfo{person}{Jingren Zhou}.} \bibinfo{year}{2019}\natexlab{}.
\newblock \showarticletitle{Hop-constrained st Simple Path Enumeration: Towards Bridging Theory and Practice.}
\newblock \bibinfo{journal}{\emph{Proc. VLDB Endow.}} \bibinfo{volume}{13}, \bibinfo{number}{4} (\bibinfo{year}{2019}), \bibinfo{pages}{463--476}.
\newblock


\bibitem[Robbins(2023)]%
        {BloodHound}
\bibfield{author}{\bibinfo{person}{Andy Robbins}.} \bibinfo{year}{2023}\natexlab{}.
\newblock \bibinfo{title}{“Bloodhound: Six Degrees of domain admin}.
\newblock \bibinfo{howpublished}{\url{https://github.com/BloodHoundAD/BloodHound}}.
\newblock
\newblock
\shownote{Accessed: 2022-08-02}.


\bibitem[Schulman et~al\mbox{.}(2017)]%
        {schulman2017proximal}
\bibfield{author}{\bibinfo{person}{John Schulman}, \bibinfo{person}{Filip Wolski}, \bibinfo{person}{Prafulla Dhariwal}, \bibinfo{person}{Alec Radford}, {and} \bibinfo{person}{Oleg Klimov}.} \bibinfo{year}{2017}\natexlab{}.
\newblock \showarticletitle{Proximal policy optimization algorithms}.
\newblock \bibinfo{journal}{\emph{arXiv preprint arXiv:1707.06347}} (\bibinfo{year}{2017}).
\newblock


\bibitem[Short and Domagalski(2013)]%
        {short2013iron}
\bibfield{author}{\bibinfo{person}{Matthew~W Short} {and} \bibinfo{person}{Jason~E Domagalski}.} \bibinfo{year}{2013}\natexlab{}.
\newblock \showarticletitle{Iron deficiency anemia: evaluation and management}.
\newblock \bibinfo{journal}{\emph{American family physician}} \bibinfo{volume}{87}, \bibinfo{number}{2} (\bibinfo{year}{2013}), \bibinfo{pages}{98--104}.
\newblock


\bibitem[Sun et~al\mbox{.}(1996)]%
        {sun1996hill}
\bibfield{author}{\bibinfo{person}{Xiaorong Sun}, \bibinfo{person}{Steve~Y Chiu}, {and} \bibinfo{person}{Louis~Anthony Cox}.} \bibinfo{year}{1996}\natexlab{}.
\newblock \showarticletitle{A hill-climbing approach for optimizing classification trees}.
\newblock In \bibinfo{booktitle}{\emph{Learning from Data: Artificial Intelligence and Statistics V}}. \bibinfo{publisher}{Springer}, \bibinfo{pages}{109--117}.
\newblock


\bibitem[{\"U}nl{\"u}yurt(2004)]%
        {unluyurt2004sequential}
\bibfield{author}{\bibinfo{person}{Tongu{\c{c}} {\"U}nl{\"u}yurt}.} \bibinfo{year}{2004}\natexlab{}.
\newblock \showarticletitle{Sequential testing of complex systems: a review}.
\newblock \bibinfo{journal}{\emph{Discrete Applied Mathematics}} \bibinfo{volume}{142}, \bibinfo{number}{1-3} (\bibinfo{year}{2004}), \bibinfo{pages}{189--205}.
\newblock


\bibitem[Vazarkar(2019)]%
        {SharpHound}
\bibfield{author}{\bibinfo{person}{Rohan Vazarkar}.} \bibinfo{year}{2019}\natexlab{}.
\newblock \bibinfo{title}{SharpHound - C\# Rewrite of the BloodHound Ingestor}.
\newblock \bibinfo{howpublished}{\url{https://github.com/BloodHoundAD/SharpHound3}}.
\newblock


\bibitem[Yu et~al\mbox{.}(2023)]%
        {yu2023deep}
\bibfield{author}{\bibinfo{person}{Zheng Yu}, \bibinfo{person}{Yikuan Li}, \bibinfo{person}{Joseph Kim}, \bibinfo{person}{Kaixuan Huang}, \bibinfo{person}{Yuan Luo}, {and} \bibinfo{person}{Mengdi Wang}.} \bibinfo{year}{2023}\natexlab{}.
\newblock \showarticletitle{Deep reinforcement learning for cost-effective medical diagnosis}.
\newblock \bibinfo{journal}{\emph{arXiv preprint arXiv:2302.10261}} (\bibinfo{year}{2023}).
\newblock


\bibitem[Zhang et~al\mbox{.}(2023)]%
        {zhang2023oracle}
\bibfield{author}{\bibinfo{person}{Yumeng Zhang}, \bibinfo{person}{Max Ward}, \bibinfo{person}{Mingyu Guo}, {and} \bibinfo{person}{Hung Nguyen}.} \bibinfo{year}{2023}\natexlab{}.
\newblock \showarticletitle{A Scalable Double Oracle Algorithm for Hardening Large Active Directory Systems}.
\newblock \bibinfo{journal}{\emph{The 18th ACM ASIA Conference on Computer and Communications Security (ACM ASIACCS)}} (\bibinfo{year}{2023}).
\newblock


\bibitem[Zhang et~al\mbox{.}(2024)]%
        {zhang2024practical}
\bibfield{author}{\bibinfo{person}{Yumeng Zhang}, \bibinfo{person}{Max Ward}, {and} \bibinfo{person}{Hung Nguyen}.} \bibinfo{year}{2024}\natexlab{}.
\newblock \showarticletitle{Practical Anytime Algorithms for Judicious Partitioning of Active Directory Attack Graphs}. In \bibinfo{booktitle}{\emph{33rd International Joint Conference on Artificial Intelligence}}. International Joint Conferences on Artificial Intelligence, \bibinfo{pages}{7074--7081}.
\newblock


\bibitem[Zheng et~al\mbox{.}(2011)]%
        {zheng2011active}
\bibfield{author}{\bibinfo{person}{Alice~X Zheng}, \bibinfo{person}{John Dunagan}, {and} \bibinfo{person}{Ashish Kapoor}.} \bibinfo{year}{2011}\natexlab{}.
\newblock \showarticletitle{Active graph reachability reduction for network security and software engineering}. In \bibinfo{booktitle}{\emph{IJCAI Proceedings-International Joint Conference on Artificial Intelligence}}, Vol.~\bibinfo{volume}{22}. \bibinfo{pages}{1750}.
\newblock


\end{thebibliography}

\section{Appendix}
\subsection{Proof of theorem \ref{theorem:np}}
\label{sec:hardness}

\begin{theorem} The APR problem is $\mathcal{\#P}$-hard. 
\label{theorem:np}
\end{theorem}

\begin{proof}

The proof is based on a reduction from the $(s, t)$- network reliability problem \cite{reliability} which is $\#\mathcal{P}$-hard.

\textbf{PROBLEM: } $(s, t)$-Network Reliability Problem 
\begin{itemize}
    \item \textbf{Input:} A graph $G = (V, E)$, source node $s$ and destination node $t$, probability $p_e \in \left[0, 1\right]$ associated with the present of each edge.
    \item \textbf{Question:} What is the probability that there is a path between two distinguished vertices s and t
\end{itemize}

First, let $Rel(G)$ denote the $(s-t)$-reliability of graph $G$, i.e., the probability that there exists a path between nodes $s$ and $t$. We assume that each edge in $G$ is operational (i.e., "On") with a probability of $p_e = 0.5$. We known that $Rel(G)$ is $\#\mathcal{P}$-hard to compute. 
Now, we proceed with the construction of the reduction instance. Suppose there exists a directed graph $G' = (V', E')$ and a query limit of $B$. Let $m = |E'|$ represent the number of edges in $G'$ and $B > m$. 
For this construction, we define two types of edges: high-confidence edges (which are more likely to be misconfigurations) and low-confidence edges (which are less likely to be misconfigurations). In this way, we assume that when a path involves $x$ high-confidence edges ($x \geq 1$), then the IT admin will never pick one of the low-confidence edge and they will only pick one of the high-confidence edges with equal probability ($1/x$). Conversely, if a path involves only low-confidence edges, the IT administrator will have to choose one of these edges with equal probability. This essentially assumes that the confidence score of high-confidence edge is infinitely. As defined in the APR problem, the IT administrator behaviour is random, i.e., they choose an edge from the proposed path to remove according to the probability distribution defined by the Equation 1 

Given a APR instance graph $G'$, we construct the following graph $G$: For each edge $(u, v)$ in $G'$, we introduce an auxiliary node called $uv$. We then add one low-confidence edge between  $u$ and $uv$ and one low-confidence edge between $uv$ and $v$. Additionally, we introduce $B$ parallel high-confidence edges between $uv$ and $v$. We refer to this constructed graph as $G$.  In this graph, for the segment $u->uv->v$, we classify the edges as follows: an edge is called an $m_l$-type edge if it connects $u$ to $uv$; an edge is called an $n_l$-type if it is a low-confidence edge connecting $uv$ to $v$; and an edge is called $n_h$-type edge if it is a high-confidence edge connecting $uv$ to $v$

The core idea behind this construction is to ensure that the optimal policy avoids presenting any path with $n_h$-type edges to the IT administrator. Querying path contain $n_h$-type edges is suboptimal because resolving any scenario will require at least $B + 1$.

The overall idea of the construction above is to ensure that the optimal policy avoids presenting any high-confidence edges to the IT administrator, since query high-confidence edges will be never useful as any situation will required to proposed at least $B+1$ query to cut the graph. Therefore, the optimal policy prioritizes querying paths that only contain $m_l$-type and $n_l$-type. 

Now, consider a single segment $u\xrightarrow[]{} uv \xrightarrow[]{} v$, which involves two low-confidence edges and $B$ high-confidence edges, being presented to the IT administrator. Following the optimal policy, $m_l$-type edge and $n_l$-edge of the segment will always be presented first. There is 50\% that the IT admin will choose an $m_l$-type edge and a 50\% chance they will choose an $n_l$-type edge. If the IT admin selects the $m_l$-type edge, the segment will be successfully disconnected (i.e., $u$ will be disconnected from  $v$ within this segment). However, if the IT admin chooses the $n_l$-type edge, the segment becomes impossible to disconnect with $B$ query. This is because once the $n_l$-type edge is removed, all $B$ parallel $n_h$-type edges must be queried to disconnect the segment (we will always hit the query limit $B$). 

Now, we will think about optimal query policy on $G$. We denote tuple $\mathcal{I} = \langle G, B \rangle$ as the APR problem instance on graph $G$ with budget of $B$
Next, let us introduce the concept of a\textit{ realization}. In general terms, a \textit{realization} is a specific outcome or instance of a random variable or process—essentially, the actual occurrence of a particular event within a probabilistic framework. In our problem context, a realization represents the decision made by the IT administrator when a path is proposed. We denote \textit{realization} with a function $\psi : P \mapsto E$. The function $\psi (p)$ acts as an oracle that returns the edge $e \in p$ that will be removed when the path $p$ is proposed to the IT administrator under realization $\psi$

Next, let $q(\mathcal{I}, \pi, \psi)$ represent the number of queries made when following the query policy $\pi$ for the problem instance $\mathcal{I}$ under the realization $\psi$. We denote $\psi_{\mathcal{I}, \pi, b>x}$ as the set of realizations where the query cost is at most $x$ when applying policy $\pi$ to instance $\mathcal{I}$. The expected number of queries across all realizations for instance $\mathcal{I}$ under policy $\pi$ is denoted by $\mathbb{E} \left[ q(\mathcal{I}, \pi) \right]$.
Additionally, the conditional expected number of queries, given that the query cost is at most $\mathbb{E} \left[ q(\mathcal{I}, \pi, \Psi) | \Psi \in \psi_{\mathcal{I}, \pi, b>x} \right]$. We define $\pi^*$ as the optimal policy.
According to the law of total expectation, the following equation holds:

\begin{equation}
\begin{split}
    \mathbb{E} &\left[ q(\mathcal{I}, \pi^*) \right] \\
    &= \mathbb{E} \left[ q(\mathcal{I}, \pi^*, \Psi) | \Psi \in \psi_{\mathcal{I}, \pi^*, b<m}\right] \cdot Pr(\Psi \in \psi_{\mathcal{I}, \pi^*, b<x}) \\
    &+ \mathbb{E} \left[ q^*(\mathcal{I}, \pi, \Psi) | \Psi \in \psi_{\mathcal{I}, \pi, b>m}\right] \cdot Pr(\Psi \in \psi_{\mathcal{I}, \pi, b>x})
\end{split}
\end{equation}

We observe that for realizations where the query cost is at most $m$ (i.e. $\psi_{\mathcal{I}, \pi^*, b<m}$), we will also successfully disconnect $(s, t)$ in graph $G$, Similarly, for realizations where the query cost exceeds $m$ (i.e. $\psi_{\mathcal{I}, \pi^*, b>m}$), we fail to disconnect $(s, t)$ in $G$. This is because there are only $m$ edges of $m_{l}$-type, and exceeding $m$ queries will always deplete all of the budget (reminding that successful disconnection of an segment requires the IT administrator to select an $m_{l}$-type). Therefore, if we define $\psi_{\mathcal{I}, \pi, s\leftrightarrow t}$ as the set of realizations where $(s,t)$ remains connected after applying the optimal query process to instance $\mathcal{I}$, and $\psi_{\mathcal{I}, \pi, s \not \leftrightarrow t}$ as the set of realizations where $(s,t)$ becomes disconnected, we have $\psi_{\mathcal{I}, \pi^*, b\leq m} = \psi_{\mathcal{I}, \pi^*, s \leftrightarrow t}$ and $\psi_{\mathcal{I}, \pi^*, b>m} = \psi_{\mathcal{I}, \pi^*, s \not \leftrightarrow t}$. This allows us to express the expected number of queries using the following equation:

\begin{equation}
\begin{split}
    &\mathbb{E} \left[ q(\mathcal{I}, \pi^*) \right] = \mathbb{E} \left[ q(\mathcal{I}, \pi^*, \Psi) | \Psi \in \psi_{\mathcal{I}, \pi^*, s \leftrightarrow t}\right] \cdot Pr(\Psi \in \psi_{\mathcal{I}, \pi^*, s \leftrightarrow t}) \\ 
    & + \mathbb{E} \left[ q^*(\mathcal{I}, \pi, \Psi) | \Psi \in \psi_{\mathcal{I}, \pi, s \not \leftrightarrow t}\right] \cdot  Pr(\Psi \in \psi_{\mathcal{I}, \pi, s \not \leftrightarrow t}) \\
    &= \mathbb{E} \left[ q(\mathcal{I}, \pi^*, \Psi) | \Psi \in \psi_{\mathcal{I}, \pi^*, s \leftrightarrow t}\right] \cdot  Pr(\Psi \in \psi_{\mathcal{I}, \pi^*, s \leftrightarrow t}) + B \cdot Pr(\Psi \in \psi_{\mathcal{I}, \pi, s \not \leftrightarrow t}) \\
    &= \mathbb{E} \left[ q(\mathcal{I}, \pi^*, \Psi) | \Psi \in \psi_{\mathcal{I}, \pi^*, s \leftrightarrow t}\right] (1 - Rel(G)) + B \cdot Rel(G) 
\end{split}
\end{equation}

From this equation, we observe that the optimal query strategy is independent of $B$ (since the second term is not controlled by the policy). Calculate the optimal query strategy is all about minimizing the expected number of queries, conditional on $G$ being $s-t$ disconnected (i.e.$\mathbb{E} \left[ q(\mathcal{I}, \pi^*, \Psi) | \Psi \in \psi_{\mathcal{I}, \pi^*, s \leftrightarrow t}\right]$. When $G$ is $s-t$ connected, then we will always hit the budget limit. We can calculate $Rel(G)$ by taking the difference between $\mathbb{E} \left[ q(\mathcal{I'}=\langle G, B+1 \rangle, \pi^*) \right]$ and $\mathbb{E} \left[ q(\mathcal{I}=\langle G, B \rangle, \pi^*) \right]$. This shows that calculating the expected number of queries in the APR problem is as difficult as the reliability problem. 

\end{proof}



    

\subsection{A note about Strong Adaptive Monotonicity and Strong Adaptive Submodular}
\label{apd:asub}

To define strong adaptive submodularity, we first need the following extension of $\Delta(a\mid s)$
\begin{definition}
    \textbf{(Conditional expected marginal benefit (extended version)} Given a state $s$ and $s'$ where $\psi_s \subseteq \psi_{s'}$, an action $a$ and a utility function $g$, the expected marginal benefit of $a$ is defined as $\Delta (a | s;s') = \sum_{e \in a}  \{ \Phi (e\mid s',a) * \left[g(s\setminus e) - g(s) \right] \}$
\end{definition}

\begin{definition}
    \textbf{(Strong Adaptive Monotonicity)} A utility function $g : S \mapsto \mathbb{R}_{\geq 0}$ is adaptive monotone if the \textit{benefit of selecting an action is always nonnegative}. Formally, function $g$ is adaptive monotonic if for all $s\in S$ and $e \in \{e \mid e \in a, a \in A_s \}$, we have $g(s) - g(s\setminus e) \geq 0$
\end{definition}

\begin{definition}
    \textbf{(Strong Adaptive Submodular\footnotemark[1])} A utility function $g : S \mapsto \mathbb{R}_{\geq 0}$ is adaptive submodular if marginal benefit of selecting an action not increase as more action are selected. Formally, function $g$ is adaptive submodular if for all temporary state $s$ and $s'$ such that $\phi_{s} \subseteq \phi_{s'}$ and action $a \in A_{s_r}$, we have $\Delta(a | s;s') \geq \Delta(a \mid s')$
\end{definition}

We also provide a the definition of pointwise submodularity as follows:

\begin{definition}
    \textbf{(Pointwise Submodular)} A utility function $g : S \mapsto \mathbb{R}_{\geq 0}$ A function $g$ say to be pointwise submodular if $g$ is submodular in every state for any realization $\psi$. Formally, function $g$ is pointwise submodular if for all temporary state $s$ and $s'$ such that $\phi_{s} \subseteq \phi_{s'}$ and for all $e \in \{e \mid e \in a, a \in A_s \}$, we have $g(s\setminus e) - g(s) \geq g(s'\setminus e) - g(s')$
\end{definition}

As our definition of adaptive monotonicity is already satisfy strong adaptive monotonicity condition so we admit the proof.

A sufficient condition for strong adaptive submodularity is that the function $g$ is both \textit{adaptive submodular} and \textit{pointwise submodular}. 
The utility function $g$ is pointwise submodular as for every state $s$ and $s'$ such that $s\subseteq s'$, and every action $a \in A_{s_r}$, we have $g(s\setminus a) - g(s) \geq g(s'\setminus a) - g(s')$. By definition, if a function $g$ is both adaptive submodular and pointwise submodular, then $g$ is strongly adaptive submodular.
        

\subsection{Pseudoscope for MDP-based Exact Algorithm}
\begin{algorithm}[H]
 \caption{Dynamic Programming (OPT)}
 \label{alg:dp}
 \begin{algorithmic}[1]
 \renewcommand{\algorithmicrequire}{\textbf{Input:}}
 \REQUIRE Directed Attack Graph $G(V, E)$, 
 \renewcommand{\algorithmicrequire}{\textbf{Output:}}
 \REQUIRE Optimal query policy $\pi$  \\
 \STATE \textbf{for} $i \in \left[ 0, B\right]$
 \STATE \quad \textbf{for} $\textbf{s} \in \mathcal{S}_i$
 \STATE \quad \quad \textbf{if} $\textbf{s}$ is in $T_d$
 \STATE \quad \quad \quad $u_{\pi}(\textbf{s}\setminus e) = 0$, $\pi(\textbf{s}\setminus e) = \emptyset$
 \STATE \quad \quad \textbf{elif} $\textbf{s}$ is in $T_b$
 \STATE \quad \quad \quad $u_{\pi}(\textbf{s}\setminus e) = \alpha$, $\pi(\textbf{s}\setminus e) = \emptyset$
 \STATE \quad \quad \textbf{else}
 \STATE \quad \quad \quad $a^* = \arg\min_{p \in A_{\textbf{s}}} \{ \sum_{e \in p}\left[  \Phi(e|s,a) * u_{\pi}(\textbf{s}\setminus e)\right]\}$
  \STATE \quad \quad \quad $u_{\pi}(\textbf{s}) = 1 + \sum_{e \in p}\left[   \Phi (e|s,a) * u_{\pi}(\textbf{s}\setminus e)\right]$
 \STATE \quad \quad \quad $\pi(\textbf{s}) = p^*$
 \STATE \textbf{return} $\pi$
 \end{algorithmic}
\end{algorithm}

\subsection{Time complexity analysis of Dynamic Programming Exact Algorithm}

According to Lemma 2, we can enumerate the action space $A_{s_r}$ in advance and the action space $A_s$ of any state $s$ can be obtained by removing any path in $A_{s_r}$ that contain edges that have been removed by IT admin during the process. Enumerating action is equal to enumerating $s-t$ path in the graph which takes $|V|^{k}$ under the DFS-based approach \cite{peng2019hop}.
The number of subproblem can grow up to $|E|^B$, running Depth First Search procedure to check if $s, t$ connectedness in every subproblem will take $\mathcal{O}(|E|+|V|)$.
Hence, the overall Exact algorithm will take us $\mathcal{O}(|V|^{k}+(|E|+|V|)*|E|^B)$. And thank to lemma 2, we avoid the complexity $\mathcal{O}(|V|^{k}*(|E|+|V|)*|E|^B)$

\subsection{Time complexity analysis of DPR algorithm}

The idea of DPR is to run the dynamic program with the restricted number of subproblem. Here we restricted the number of subproblem to $\mathcal{O}(\tau)^{B'}$. This make the complexity of DPR become $\mathcal{O}(|V|^{k}+(|E|+|V|)*(\tau)^{B'})$

\subsection{Vulnerability level and the number of cross-tier misconfig. in the attack graph.}

\begin{table}[H]
\begin{center}
\caption{Vulnerability level and the number of cross-tier misconfig. in the attack graph.}\label{tab:misconfig}
\smallskip\noindent
\resizebox{0.5\linewidth}{!}{%
\begin{tabular}{cccccc}
\hline
                                & level 1   & level 2 & level 3 & level 4 & level 5 \\
\hline
\textbf{$\#$ of misconfig.}     & 9         & 27      & 45      & 98      & 197                             \\
\hline
\end{tabular}}
\end{center}

\end{table}

\subsection{Proof of theorem 2}
\begin{theorem}
\label{theorem:approx1}
 Algorithm 1 achieves a $(\ln{|P_{s_r}| + 1)^2}$-approximation for the APR problem with $B = |P_{s_r}|$.    
\end{theorem}

\begin{proof}
    Applying Theorem 17 from \cite{golovin2011adaptive}, the approximation ratio for the greedy algorithm, which maximizes marginal gain, is bounded by:
    \begin{equation}
        c(\pi) \leq \alpha c(\pi^*)\left(\ln{\frac{Q}{\eta}} + 1\right)^2
    \end{equation}
    where $c(\pi)$ is the average cost of the greedy policy which is $\alpha$-approximate w.r.t items cost, $c(\pi^*)$ is the cost of the optimal policy, $Q$ is the utility target for covering the ground set $U$, $\eta$ is the threshold parameter such that $f(S) > Q - \eta$ will implies $f(S) = Q$.

    For algorithm 1 and utility function $g$, we have $range(g) \subseteq \mathbb{N}$ which imply that $\eta = 1$. In the context of the APR problem, achieving full coverage requires to disconnect every path in $P_{s_r}$. This condition is satisfied when $Q = |P_{s_r}|$. Also, $\alpha = 1$ for algorithm 1. Therefore, the following approximation ratio holds for Algorithm 1:
    \begin{equation}
        q(\pi) \leq q(\pi^*)\left(\ln{|P_{s_r}|} + 1\right)^2
    \end{equation}
\end{proof}
\subsection{Other Algorithm}

\subsubsection{Reinforcement Learning}
\label{sec:rl}
In this approach, we adopt a strategy similar to the \textit{DPR} algorithm by limiting the action space to make the problem more manageable. However, instead of solving the problem optimally using Dynamic Programming, we employ a model-free reinforcement learning approach to approximate learn the query policy. 
We utilize two model-free Actor-Critic reinforcement learning models: Proximal Policy Optimization (PPO) \cite{schulman2017proximal} and Soft Actor-Critic for Discrete Action (SAC) \cite{christodoulou2019soft}. These Actor-Critic methods allow us to train a reinforcement learning agent across multiple environments simultaneously, where each environment represents a possible realization $\psi$ of the APR problem. The objective in the APR problem is to derive a policy that minimizes the expected number of queries across all possible realizations. By interacting with several realizations, the RL agent learns a policy that minimizes the overall reward (corresponding to the number of queries) for these scenarios.
For the implementation of RL, we encode the observation space as a vector of $(E + \tau B)$ binary bits. The first $E$ bits represent a one-hot encoding of the edges that have been removed through queries, while the remaining $\tau B$ bits encode the actions that have been taken.

\subsubsection{Other Heuristic} 
\label{sec:oheur}

In this section, we introduce two additional heuristics that performed surprisingly well compared to the APP algorithm.

\textbf{Minimum-cut-based heuristic (OTH1)}
The idea of this heuristic is to proposes paths with the highest likelihood of removing an edge in the minimum cut set. Formally, it selects the path $p = \arg\max_{a \in P_{G'}:a \cap mc(G')} \sum_{e \in a}  \{ \Phi (e|s,a) * \left[g(s\setminus e) - g(s) \right] \}$ where $mc(G')$ return the $s-t$ minimum cut of the temporary graph $G'$. Suppringingly, MC is actually an approximate algorithm  with the approximation ratio of $|P_{s_r}|$ when queries have a cost (i.e., $c > 1$). However, since our problem assumes unit cost (i.e., $c = 1$), this approximation bound becomes trivial, and we treat MC as a heuristic. Despite this, the MC heuristic outperforms the APP algorithm in 9 out of 12 graphs in our experiments, suggesting the potential for a tighter approximation bound and leave this for future research.

\textbf{Approximate Algorithm on Shortest Paths (OTH2)} In this heuristic, instead of running the APP algorithm on all simple paths, we restrict it to only the set of shortest paths. To remind, the APP algorithm is used to find the path that yields the highest marginal gain within the given set of paths. Surprisingly, this simple modification performs exceptionally well, outperforming the original APP algorithm in 10 out of 12 graphs.


\subsection{More about sampling scheme for DPR and RL algorithm}

\textbf{Sampling Scheme for DPR and RLs.} Now we will discuss about the sampling scheme for DPR, and both of the RL technique.  In total, we have 3 sampling scheme for each algorithm. We implement our sampling algorithm by modifying the APP, OTH1 and OTH2 algorithm to sample the action. The idea is instead of return the top action proposed by these algorithm, we return the top-k path by run these algorithm. The idea is to use DPR or RLs techniques to find the best action among $k$ proposed paths in each step. 
Specifically, the APPwill propose the path with the highest marginal gain with respect to utility $g$, here, we use APPto return top $k$ path with highest marginal gain. Similarly, for OTH2, we return top $k$ shortest path. For OTH1, we return top $k$ path that have the highest likelihood of removing a edge in minimum cut. In our experiment, we fix $k=4$ for DPR and $k=16$ for the reinforcement learning methods. This choice is due to DPR's limited scalability at $k=4$, while PPO and SAC can scale effectively to $k=16$.  In the experiment, for each of algorithm, we will report the result of the sampling scheme that yield the highest performance for each algorithm.

\subsection{Data collection and preprocessing for the ORG graph}
The data was collected from an anonymous organization using SharpHound ~\cite{SharpHound}, which gathers detailed information on user sessions, group memberships, ACLs (Access Control Lists), and permissions within the AD environment. SharpHound is commonly used tool to collect domain data, which is then imported into BloodHound for visualizing potential attack paths. The data collection took place on Thursday, 28th October 2022, at approximately 10:00 AM during regular working hours in the local time zone. The collected Active Directory dataset consists of 125,444 nodes and 1,195,432 edges. While we do not have specific insight into the number of tiers in this AD instance, nor the exact rules the organization uses to define them (due to the confidentiality), we assume the AD follows the common three-tier model. We heuristically classify nodes into three tiers: nodes flagged as "adminaccount" (which SharpHound identifies as highly privileged groups) are categorized as Tier 0, servers and services are placed in Tier 1 (determined by their name), and all remaining nodes are assigned to Tier 2. In this experiment, we assumed the  undefined-tier ratio of 0.95. Given the shear size of this AD graph, we increased the budget from 10 to $B = 20$ and $B = 30$ queries. Additionally, due to runtime constraints, we were unable to include results for the DPR algorithm, as generating queries for 16,000 episodes for this algorithm would take approximately 800 seconds per episode.

\subsection{On the length of the proposed path}

The length of the proposed path is an important consideration in the proposing process, as longer paths could place additional burden on the IT administrator. Table \ref{tab:len} reports the expected average lengths of the proposed paths in our experiments. The results show that the proposed paths are relatively short, even for real AD environments (DPR average path length is only 2.776 for ORG). This indicates that our approach does not create significant burdens for IT administrators.

\begin{table*}
\begin{center}
\caption{Average length of the queried path by algorithm. ($\downarrow$ is better) }\label{tab:len}
\smallskip\noindent
\resizebox{1\linewidth}{!}{%
\begin{tabular}{cccccccccccc}
\toprule
        ~      & \textbf{G1}& \textbf{G2} & \textbf{G3} & \textbf{G4} & \textbf{G5} & \textbf{G6} & \textbf{G7} & \textbf{G8} & \textbf{G9} & \textbf{ORG}(20)& \textbf{ORG}(30)  \\\hline
        APP& 2.228           & 2.219            & 2.524            & 2.178            & 2.252            & 2.140            & 2.864             & 1.862             & 2.162            & 3.488          & 3.400   \\ \hline
        OTH1 & 2.228           & 2.218         & 2.473           & 1.923         & 2.252            & 2.139            & 2.864            & 1.863             & 2.162            & 2.769           & 2.809        \\ \hline
        OTH2 &  2.258            & 2.238          & 2.476           & 1.894           & 2.279            & 2.159             & 2.902             & 1.861            & 2.184           & 2.724          & 2.778  \\ \hline
        PPO    & 2.259           & 2.236        & 2.474            & 1.893            & 2.279            & 2.158            & 2.901             & 1.862             & 2.185            & 2.726           & 2.788    \\ \hline  
        SAC    & 2.259           & 2.238            & 2.477            & 1.894            & 2.279   & 2.158            & 2.907           & 1.861            & 2.186            &  2.728          & 2.783      \\ \hline
        DPR   & 2.259          & 2.237   & 2.473   & 1.894            & 2.277            & 2.159   & 2.901             & 1.86    & 2.184   &  2.72          & 2.776         \\ 
\toprule
\end{tabular}}
\end{center}

\end{table*}

\end{document}